\documentclass{article}

\usepackage{microtype}
\usepackage{graphicx}
\usepackage{booktabs} % for professional tables

\usepackage{hyperref}
\usepackage{adjustbox}
\usepackage{subcaption}
\usepackage{multirow}
\usepackage{bigstrut}

\usepackage[accepted]{icml2025}

\usepackage{amsmath}
\usepackage{amssymb}
\usepackage{mathtools}
\usepackage{amsthm}

\usepackage{colortbl} % For coloring rows
\definecolor{color3}{gray}{0.95}
\definecolor{color4}{rgb}{0.95, 0.91, 0.91}
\definecolor{color5}{rgb}{0.35, 0.71, 0.67}
\definecolor{color6}{rgb}{0.17, 0.48, 0.71}

\def\eg{e.g.}

\usepackage[capitalize,noabbrev]{cleveref}

\theoremstyle{plain}

\theoremstyle{definition}

\theoremstyle{remark}

\usepackage[textsize=tiny]{todonotes}

\icmltitlerunning{}
\begin{document}

\twocolumn[
\icmltitle{BinaryHPE: 3D Human Pose and Shape Estimation via Binarization}



\icmlsetsymbol{equal}{*}

\begin{icmlauthorlist}
\icmlauthor{Zhiteng Li}{yyy}
\icmlauthor{Yulun Zhang}{yyy}
\icmlauthor{Jing Lin}{yyy2}
\icmlauthor{Haotong Qin}{sch}
\icmlauthor{Jinjin Gu}{yyy3}
\icmlauthor{Xin Yuan}{yyy4}
\icmlauthor{Linghe Kong}{yyy}
\icmlauthor{Xiaokang Yang}{yyy}
\end{icmlauthorlist}

\icmlaffiliation{yyy}{Shanghai Jiao Tong University}
\icmlaffiliation{yyy2}{Tsinghua University}
\icmlaffiliation{yyy3}{The University of Sydney}
\icmlaffiliation{yyy4}{Westlake University}
\icmlaffiliation{sch}{ETH Z\"{u}rich}

\icmlcorrespondingauthor{Yulun Zhang}{yulun100@gmail.com}
\icmlcorrespondingauthor{Linghe Kong}{linghe.kong@sjtu.edu.cn}

\icmlkeywords{Machine Learning, ICML}

\vskip 0.3in
]



\printAffiliationsAndNotice{}  % leave blank if no need to mention equal contribution


\begin{abstract}
3D human pose and shape estimation (HPE) aims to reconstruct the 3D human body, face, and hands from a single image. Although powerful deep learning models have achieved accurate estimation in this task, they require enormous memory and computational resources. Consequently, these methods can hardly be deployed on resource-limited edge devices. In this work, we propose BinaryHPE, a novel binarization method designed to estimate the 3D human body, face, and hands parameters efficiently. Specifically, we propose a novel binary backbone called Binarized Dual Residual Network (BiDRN), designed to retain as much full-precision information as possible. Furthermore, we propose the Binarized BoxNet, an efficient sub-network for predicting face and hands bounding boxes, which further reduces model redundancy. Comprehensive quantitative and qualitative experiments demonstrate the effectiveness of BinaryHPE, which has a significant improvement over state-of-the-art binarization algorithms. Moreover, our BinaryHPE achieves comparable performance with the full-precision method Hand4Whole while using only \textbf{22.1\%} parameters and \textbf{14.8\%} operations. We will release all the code and pretrained models.
\end{abstract}

\setlength{\abovedisplayskip}{2pt}
\setlength{\belowdisplayskip}{2pt}
\vspace{-4mm}
\section{Introduction}
\label{sec:intro}
\vspace{-1mm}

3D human pose and shape estimation, also known as whole-body human mesh recovery, is a fundamental task in computer vision. It aims to reconstruct the 3D mesh of a person's entire body from a single image or video. By recovering the human mesh, we are able to understand human behaviors and feelings through their poses and expressions. Therefore, 3D human pose and shape estimation has been widely applied for action recognition, virtual try-on, motion retargeting, and more. Recently, powerful deep learning models~\citep{choutas2020monocular,rong2021frankmocap,feng2021collaborative,moon2022accurate,lin2023one} have been proposed with remarkable performance.
However, real-world applications like Augmented Reality (AR) require real-time responses, demanding models that are accurate and efficient with less memory and computation cost.

\begin{figure}
% \small
\scriptsize
\centering
% \resizebox{0.50\textwidth}{!}{
% % one row
\vspace{-3mm}
\begin{adjustbox}{valign=t}
\hspace{-3.5mm}
\begin{tabular}{cccc}
\includegraphics[width=0.245\linewidth]{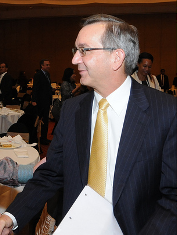} \hspace{-4mm} &
\includegraphics[width=0.245\linewidth]{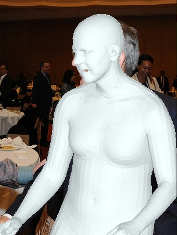} \hspace{-4mm} &
\includegraphics[width=0.245\linewidth]{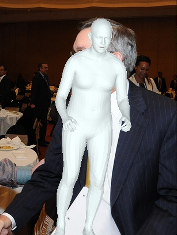} \hspace{-4mm} &
\includegraphics[width=0.245\linewidth]{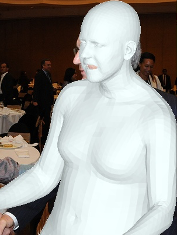} \hspace{-4mm} 
\\
% \multirow{2}{*}{Image}  \hspace{-4mm} &
Image \hspace{-4mm} &
Hand4Whole \hspace{-4mm} &
BNN \hspace{-4mm} &
\textbf{BinaryHPE (Ours)} \hspace{-4mm}
\\
Params / OPs \hspace{-4mm} & 
77.84 / 16.85  \hspace{-4mm} &
21.61 / 2.63 \hspace{-4mm} &
\textbf{17.22} / \textbf{2.50} \hspace{-4mm}
\end{tabular}
\end{adjustbox}
%}
\vspace{-2.7mm}
% \hspace{-1mm}
\caption{\small{Comparison of full-precision Hand4Whole, BNN, and BinaryHPE. The second line is Parameters (M) / Operations (G).}}
\label{fig:intro_viscomp}
\vspace{-6mm}
\end{figure}

\begin{figure*}[t]
\vspace{-1.4mm}
\centering
% \fbox{\parbox[c][9cm]{\linewidth}{Abstract}}
\includegraphics[width=1.0\textwidth]{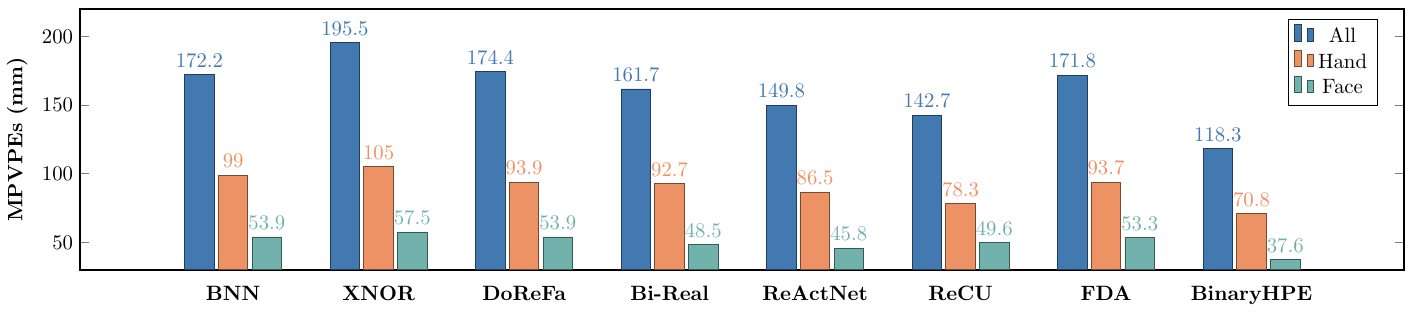}
\vspace{-8.8mm}
\caption{Comparison between recent BNNs and BinaryHPE on EHF. BinaryHPE significantly reduces the \emph{All MPVPEs} (the lower, the better) of BNN~\citep{hubara2016binarized}, XNOR~\citep{rastegari2016xnor}, DoReFa~\citep{zhou2016dorefa}, Bi-Real~\citep{liu2018bi}, ReActNet~\citep{liu2020reactnet}, ReCU~\citep{xu2021recu} and FDA~\citep{xu2021learning} by 53.9, 77.2, 56.1, 43.4, 31.5, 24.4, and 53.5 respectively.}
\label{fig:intro}
\vspace{-4.2mm}
\end{figure*}

HPE methods are generally categorized into optimization-based and regression-based approaches, with the latter gaining prominence due to the introduction of parametric models like SMPL~\citep{loper2023smpl} and SMPL-X~\citep{pavlakos2019expressive}. Most regression-based methods~\citep{choutas2020monocular,rong2021frankmocap,feng2021collaborative,moon2022accurate,zhou2021monocular} utilize separate networks for the body, hands, and face. Predicted boxes are used to crop hand and face regions from the original image, which are then resized and passed through corresponding encoders for refined estimation. Each encoder extracts high-quality features, which are decoded to regress body, hands, and face parameters. These parameters are then processed by an SMPL-X layer~\citep{pavlakos2019expressive} to reconstruct the 3D whole-body human mesh.
While these methods achieve strong performance, they are resource-intensive, often requiring large models and high-end GPUs. For example, Hand4Whole~\citep{moon2022accurate} employs a multi-stage pipeline with additional hand-only and face-only datasets~\citep{moon2020interhand2,zimmermann2019freihand}, further increasing system complexity.
As the need for efficient HPE models on mobile devices grows, there is an urgent push for simpler, resource-efficient algorithms that maintain high estimation accuracy.

As deep learning models grow rapidly in size, model compression becomes crucial, particularly for deployment on edge devices. Relevant research can be divided into five categories, including quantization~\citep{xia2022basic,qin2020forward,qin2020binary,hubara2016binarized,zhou2016dorefa,liu2018bi}, knowledge distillation~\citep{hinton2015distilling,chen2018darkrank,zagoruyko2016paying}, pruning~\citep{han2015learning,han2015deep,he2017channel}, lightweight network design~\citep{howard2017mobilenets,zhang2018shufflenet,ma2018shufflenet}, and low-rank approximation~\citep{denton2014exploiting,lebedev2014speeding,lebedev2016fast}. Among these, binarized neural network (BNN) is the most aggressive quantization technology that can compress memory and computational costs extremely. By quantizing the full-precision (32 bits) weights and activations into only 1 bit, BNN achieves significant computational efficiency, offering up to 32$\times$ memory saving and 58$\times$ speedup on CPUs for convolution layer~\citep{rastegari2016xnor}. Additionally, bitwise operations like XNOR can be efficiently implemented on embedded devices~\citep{zhang2019dabnn,ding2019regularizing}.

However, the direct application of network binarization for HPE may encounter three challenges: \textbf{(1)} The quality of extracted features from the encoder is significant for parameter regression. Directly binarizing the encoder may cause severe full-precision information loss. \textbf{(2)} The dimension mismatch problem, when reshaping features, prevents bypassing full-precision information in BNN, which should be tackled for general situations. \textbf{(3)} To obtain accurate enough body, hands, and face parameters with as little memory and computation cost as possible, which parts should or should not be binarized requires careful consideration.

To address the above challenges, we propose \textbf{BinaryHPE}, a novel BNN-based method for the HPE task.
\textbf{First}, we propose the \textbf{Bi}narized \textbf{D}ual \textbf{R}esidual \textbf{N}etwork (\textbf{BiDRN}), a novel binarized backbone featuring the Binarized Dual Residual Block (BiDRB) as its core unit. Specifically, BiDRB can bypass full-precision activations, which is significant for body, hands, and face parameter regression, by adopting a Local Convolution Residual (LCR) with almost the same memory and computation cost. Besides, we redesign four kinds of convolutional modules and generalize them to more complicated situations so they can apply the LCR even to dimension mismatch situations. Moreover, BiDRB utilizes a full-precision Block Residual (BR) to further enhance the full-precision information with tolerable cost but significant improvements. \textbf{Second}, we propose Binarized BoxNet for predicting face and hands bounding boxes, which can maintain the performance while significantly reducing memory and computation costs. Building on these two networks, we introduce BinaryHPE, which significantly outperforms state-of-the-art BNNs, achieving a reduction of over 31.5 \emph{All MPVPEs}, as shown in \Cref{fig:intro}.

Our contributions can be summarized as follows.
\vspace{-2mm}
\begin{itemize}
\vspace{-2mm}
\item We propose BinaryHPE, a novel BNN-based method designed for 3D human pose and shape estimation. To the best of our knowledge, this is the first study to explore the binarization of the HPE task.
\vspace{-2mm}
\item We propose BiDRN, a new binarized backbone with the basic unit BiDRB, consisting of Local Convolution Residual (LCR) and Block Residual (BR). This design preserves full-precision information as much as possible, narrowing the \emph{All MPVPEs} gap with the full-precision method from \textbf{85.9} to \textbf{32.0}.
\vspace{-2mm}
\item The proposed BinaryHPE not only significantly outperforms existing SOTA BNNs, but also achieves performance comparable to the full-precision Hand4Whole method, while requiring less than a quarter of the parameters and computational resources.
\end{itemize}

\begin{figure*}[t]
\vspace{-1.5mm}
\centering
% \fbox{\parbox[c][9cm]{\linewidth}{Abstract}}
\includegraphics[width=1.0\textwidth]{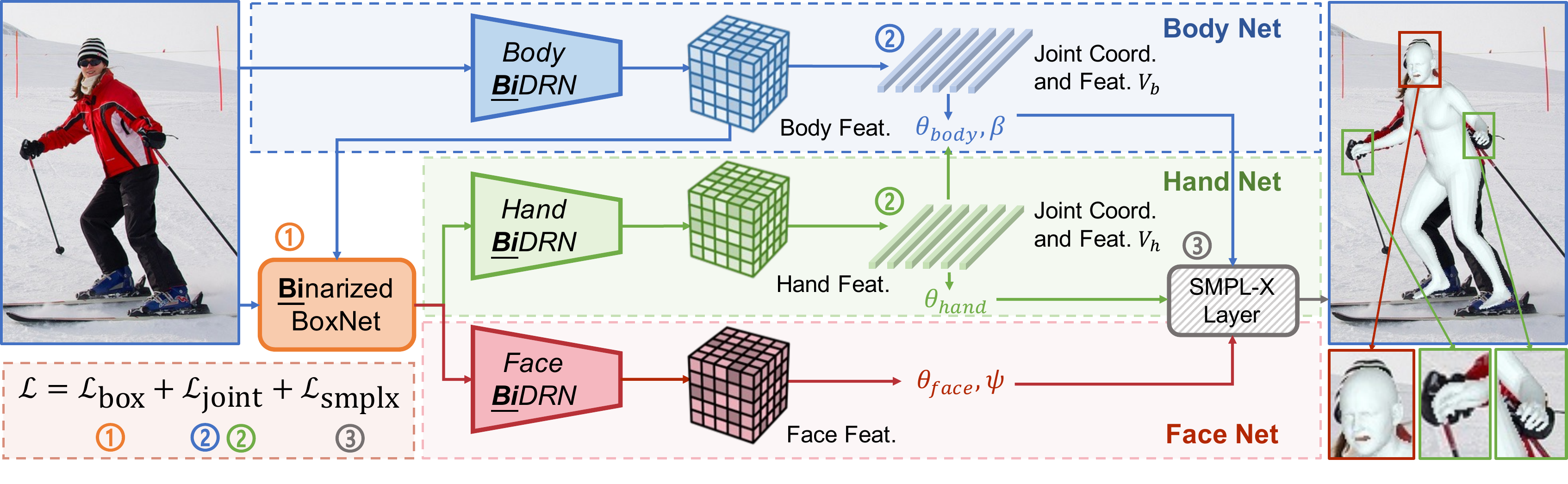}
\vspace{-10mm}
\caption{The overview pipeline of our binarized 3D human pose and shape estimation method. The body, hand, and face BiDRN serve as encoders to extract corresponding features. Binarized BoxNet predicts the face and hand regions based on the body features.}
\label{fig:arch}
\vspace{-3.5mm}
\end{figure*}

\vspace{-3mm}
\section{Related Work}
\label{sec:formatting}
\vspace{-1mm}

% \subsection{Whole-body Human Mesh Recovery}
% \vspace{-1mm}

\textbf{Human Pose and Shape Estimation.}\quad
Optimization-based methods~\citep{joo2018total,xiang2019monocular,pavlakos2019expressive,xu2020ghum} estimate 2D keypoints first and reconstruct 3D human bodies with additional constraints but are computationally intensive due to complex optimization. With the advent of statistical human models like SMPL~\citep{loper2023smpl} and SMPL-X~\citep{pavlakos2019expressive}, regression-based methods enable end-to-end 3D human mesh recovery.
For instance, ExPose~\citep{choutas2020monocular} leverages body-driven attention to refine face and hand regions. FrankMocap~\citep{rong2021frankmocap} performs independent 3D pose regression for body, face, and hands before integrating the results. PIXIE~\citep{feng2021collaborative} introduces a moderator to adaptively fuse body part features with realistic facial details. Hand4Whole~\citep{moon2022accurate} enhances 3D wrist rotation and ensures smooth body-hand transitions using both body and hand MCP joint features.
Despite their strong performance, these methods demand high memory and computational power. Their multi-stage pipelines also complicate training and increase resource consumption. However, efficient HPE models for resource-constrained devices remain largely unexplored, and this work aims to bridge that gap.

% \vspace{-1mm}
% \subsection{Binarized Neural Network}
% \vspace{-1mm}

\textbf{Binarized Neural Networks.}\quad
Binarized neural networks (BNNs)~\citep{hubara2016binarized} represent both the activations and weights with only 1-bit, providing an extreme level of compression for computation and memory. 
It is first introduced in the image classification task, and several follow-up works (\eg, Bi-Real~\citep{liu2018bi}, ReActNet~\citep{liu2020reactnet}, and IR-Net~\citep{qin2020forward}) further push the performance boundary, making substantial improvements over the original implementation. Due to BNN’s ability to achieve extreme model compression while delivering relatively acceptable performance, it has also been widely applied in other vision tasks. For example, \citet{jiang2021training} proposes a BNN without batch normalization for image super-resolution task. \citet{cai2023binarized} designed a binarized convolution unit BiSR-Conv that can adapt the density and distribution of hyperspectral image (HSI) representations for HSI restoration.
However, the potential of BNN in human pose and shape estimation remains unexplored.

\begin{figure*}[t]
% \vspace{-1mm}
\centering
  % \fbox{\parbox[c][9cm]{\linewidth}{Abstract}}
  \vspace{-3mm}
   \includegraphics[width=1.0\textwidth]{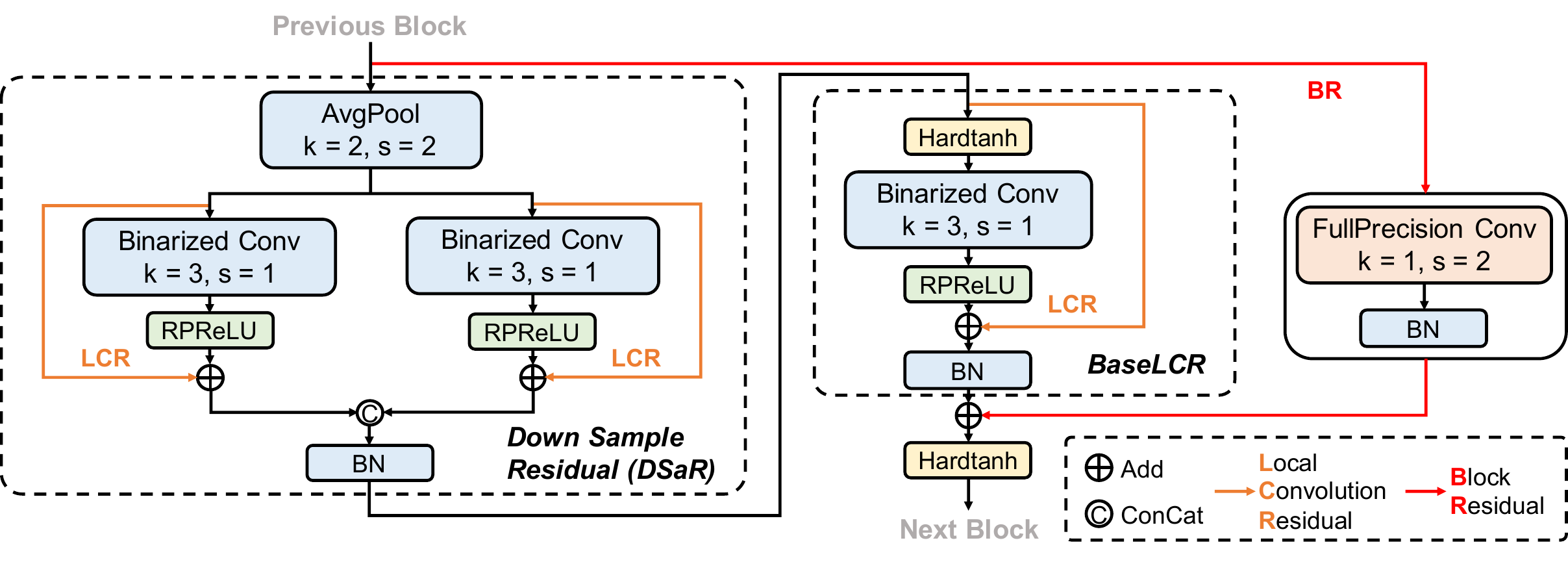}
   \vspace{-8mm}
   \caption{A Binarized Dual Residual Block (BiDRB) composed of both Local Convolution Residual (LCR) and Block Residual (BR).}
   \label{fig:BiDRB}
   \vspace{-5mm}
\end{figure*} 

\vspace{-2mm}
\section{Method}
\label{sec:method}
\vspace{-1mm}

We propose BinaryHPE, the first binary method for the HPE task, building upon the widely used Hand4Whole~\citep{moon2022accurate}. In Hand4Whole, the ResNet~\citep{he2016deep} backbones play a crucial role in extracting detailed and high-quality features from the body, face, and hands, which are the primary sources of memory and computational costs.
In addition, Hand4Whole uses the extracted body feature to predict the bounding box of face and hands by a BoxNet, which may be complex and can be compressed as well. Based on these insights, we propose a novel binary backbone, the Binarized Dual Residual Network (BiDRN) (see~\Cref{fig:arch}), to replace ResNet, along with a Binarized BoxNet. These innovations significantly reduce memory and computational costs while maintaining accuracy.

\vspace{-1mm}
\subsection{Binarized Dual Residual Block}
\vspace{-1mm}

BiDRB is the fundamental unit of BiDRN, and its details are shown in~\Cref{fig:BiDRB}. We first binarize the full-precision activation $\boldsymbol{a}_f \in \mathbb{R}^{C \times H \times W}$ to 1-bit by the Sign function:
\begin{align}
    a_b = \mathrm{Sign}(a_f) = 
    \begin{cases}
        +1, & a_f \geq 0 \\
        -1, & a_f < 0
    \end{cases},
\end{align}
where $\boldsymbol{a}_b \in \mathbb{R}^{C \times H \times W}$ denotes the binarized activation.
Yet, the Sign function is non-differentiable and we have to approximate it during backpropagation. Here, we adopt a piecewise quadratic function to smoothly approximate the Sign function during the gradient computation process as
\begin{align}
    F(a_f) =
    \left\{
    \begin{aligned}
    &+1, & a_f \geq 1 &\\
    &-a_f^2 + 2a_f,&  0\leq a_f< 1 &\\
    &a_f^2 + 2a_f, & -1 \leq a_f< 0 &\\
    &-1, & a_f < -1 &
    \end{aligned}\ .
    \right.
\end{align}
We find the ReLU pre-activation used by default in previous work will generate all-one activations after the Sign function. This may lead to the failure of binarization. To solve it, we adopt a Hardtanh pre-activation function that can compress the full-precision activation into the range $[-1,+1]$ as
\begin{align}
    a_f = \mathrm{Hardtanh}(x_f) = 
    \left\{
    \begin{aligned}
    &+1, & x_f \geq 1 &\\
    &x_f,  & -1 \leq x_f < 1 \\
    &-1, & x_f < -1 &
    \end{aligned}\ ,
    \right.
\end{align}
where $\mathbf{X}_f \in \mathbb{R}^{C \times H \times W}$ represents the output feature map generated by the preceding layer. Compared with methods that use a learnable threshold before the Sign function~\citep{liu2020reactnet} or applying a redistribution trick~\citep{cai2023binarized}, the Hardtanh pre-activation can achieve better performance without introducing additional parameter burden.

\vspace{-1mm}
Quantizing weights by the same Sign function can extremely reduce the parameters, thus weights $\mathbf{W}_f \in \mathbb{R}^{C_\text{in} \times C_\text{out} \times K \times K}$ in convolution layers are quantized into scaled 1-bit $\mathbf{W}_b$ as
\begin{align}
    w_b^i = \alpha^i \cdot \mathrm{Sign}(w_f^i),
\end{align}
where index $i$ represents the $i$-th output channel, and $\alpha^i$ is a scaling factor defined as $\alpha^i = \frac{\|w_f^i\|_1}{C_\text{in} \times K \times K}$. Multiplying the binarized weights by channel-wise scaling factor can better maintain the original distribution of weights on each channel.
After binarizing both the activations and weights, the computation of binarized convolution can be simply formulated as~\citep{rastegari2016xnor}
\begin{align}
    \boldsymbol{o} = \alpha \cdot \text{bitcount}(\mathrm{Sign}(\boldsymbol{a}_f) \odot \mathrm{Sign}(\mathbf{W}_f)),
\end{align}
where $\odot$ denotes the XNOR-bitcount bitwise operation between binarized activations and weights. 

\vspace{-1mm}
XNOR and bitcount are both logical operations that can significantly reduce the computation overhead of full-precision matrix multiplication. However, the loss of full-precision information in quantization is non-neglectable. Compared with the binarized information, full-precision information usually represents image details, which may not be dominant in the image classification task, but is significant in HPE. Since regression-based methods only optimize a few body, hands, and face parameters, even slight perturbations on the feature space may be transmitted to the parameters and have a great impact on the final 3D human mesh. 

\vspace{-1mm}
To preserve the full-precision information as much as possible, we design two kinds of residual connections, \textit{i.e.,} Local Convolution Residual (LCR) and Block Residual (BR).

\begin{figure*}[t]
  \centering
  \vspace{-3mm}
  % \fbox{\parbox[c][9cm]{\linewidth}{Abstract}}
   \includegraphics[width=1.0\textwidth]{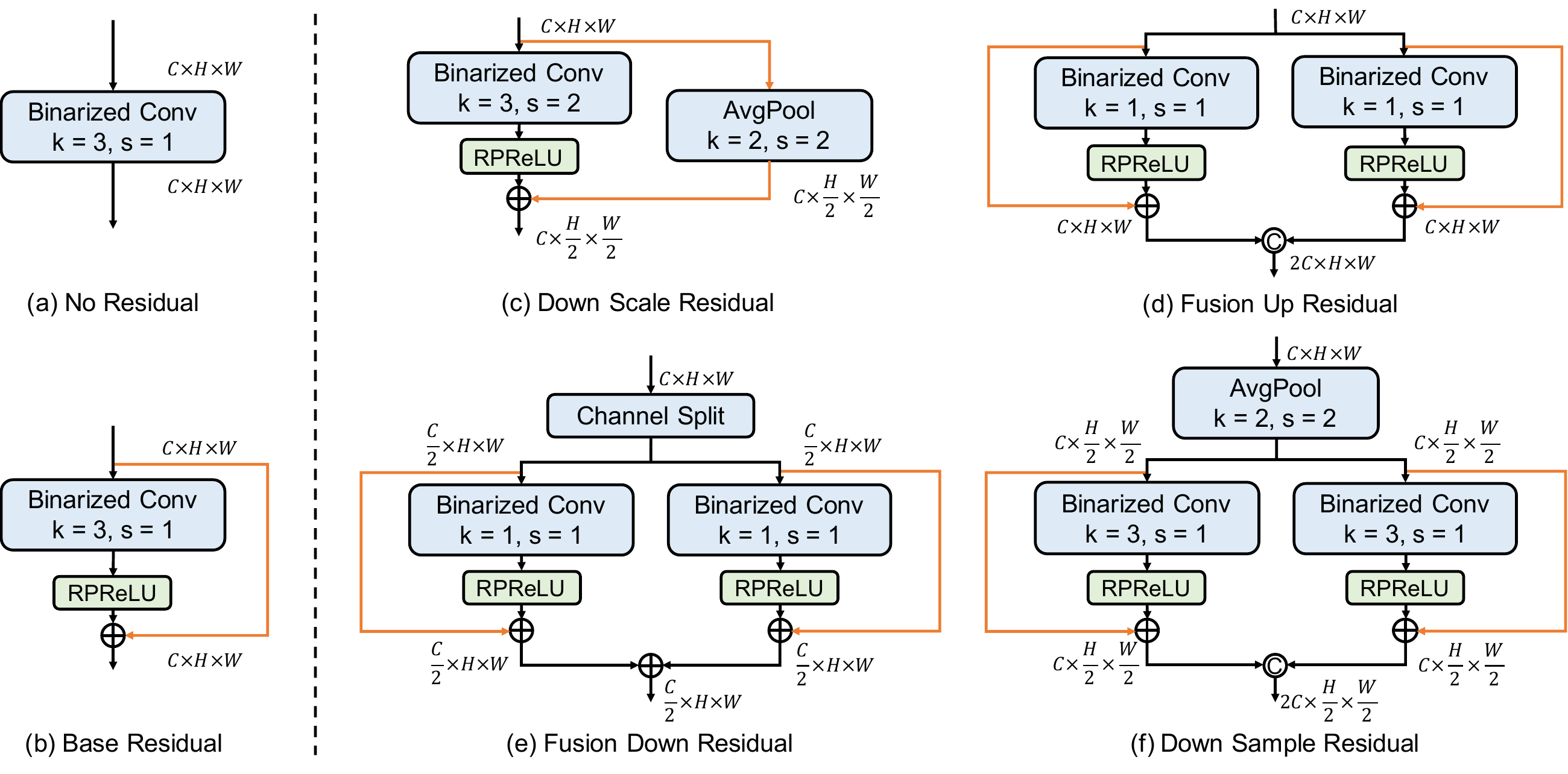}
   \vspace{-8.8mm}
   \caption{Illustration of our Local Convolution (Base) Residual and four redesign modules, including (c) Down Scale Residual (DScR), (d) Fusion Up Residual (FUR), (e) Fusion Down Residual (FDR), and (f) Down Sample Residual (DSaR). The \textcolor{orange}{orange} arrow denotes the full-precision information flow. For simplicity, batch normalization and Hardtanh pre-activation are omitted.}
   \label{fig:local_residual}
   \vspace{-4mm}
\end{figure*}

\vspace{-1mm}
\noindent\textbf{Local Convolution Residual.}\quad
This residual connection is applied to each binarized convolution layer to bypass full-precision activation. Since the value range of binarized output $\boldsymbol{o}$ is much smaller than that of full-precision activation $\boldsymbol{a}_f$, we first apply the channel-wise RPReLU~\citep{liu2020reactnet} activation function to enlarge its value diversity and redistribute the representation as
\begin{align}
    \text{RPReLU}(o^i) = 
    \left\{
    \begin{aligned}
    &o^i - \gamma^i + \zeta^i, & o^i > \gamma^i &\\
    &\beta^i(o^i - \gamma^i) + \zeta^i,  & o^i \leq \gamma^i
    \end{aligned}\ ,
    \right.
\end{align}
where $o^i$ is the binarized convolution output of the $i$-th channel, $\gamma^i, \zeta^i, \beta^i \in \mathbb{R}$ are learnable parameters. After that, the full-precision activation $\boldsymbol{a}_f$ is added as
\begin{align}
    \boldsymbol{o}' = \text{BatchNorm}(\text{RPReLU}(\boldsymbol{o}) + \boldsymbol{a}_f),
\end{align}
where $\boldsymbol{o}'$ is the output feature. Note that the parameters introduced by RPReLU are relatively small compared to the convolution kernels and thus can be ignored. 

This local convolution residual can bypass full-precision information during the whole network if the dimension remains unchanged. Unfortunately, to extract compact image features, there exists Down Scale, Down Sample, Fusion Up, and Fusion Down operations in the encoder. The dimension mismatch problem in these modules prevents bypassing the full-precision information and thus leads to a performance drop. To tackle this problem, we redesign these modules so that they can be combined with our Local Convolution Residual, as illustrated in~\Cref{fig:local_residual}. 

Specifically, Down Scale module reduces the spatial dimension of the input feature. To match the output dimension, the full-precision activation is first averaged pooled and then added to the Down Scale convolution output:
\begin{align}
    \boldsymbol{o}' = \text{BatchNorm}(\text{RPReLU}(\boldsymbol{o}) + \text{AvgPool}(\boldsymbol{a}_f)),
\end{align}
where $\boldsymbol{o}', \boldsymbol{o} \in \mathbb{R}^{C \times \frac{H}{2} \times \frac{W}{2}}, \boldsymbol{a}_f \in \mathbb{R}^{C \times H \times W}$. Average pooling does not introduce any additional parameter and its computational cost can be ignored compared to the encoder. 

For Fusion Up which increases the channel dimension, we replace the single convolution layer with two distinct layers. The design is guided by the principle of maintaining the output channel count of each layer equivalent to its input channel count. By aligning the input and output dimensions in this manner, the layers can seamlessly integrate with the normal Local Convolution Residual (LCR) mechanism, which helps in retaining the original full-precision information.
Finally, the outputs of these two layers are concatenated in channel dimension as
\begin{align}
\boldsymbol{o}' = \text{BatchNorm}(\text{Concat}(\boldsymbol{o}'_1, \boldsymbol{o}'_2)),
\end{align}
where $\boldsymbol{o}' \in \mathbb{R}^{2C \times H \times W}, \boldsymbol{o}'_1,  \boldsymbol{o}'_2 \in \mathbb{R}^{C \times H \times W}$. 

Fusion Down is the inverse of Fusion Up, thus we first split the input w.r.t. channel and then feed them into two distinct binary conv layers with LCR. Finally, they are summed as
\begin{align}
\boldsymbol{o}' = \text{BatchNorm}(\boldsymbol{o}'_1 + \boldsymbol{o}'_2),
\end{align}
where $\boldsymbol{o}' \in \mathbb{R}^{\frac{C}{2} \times H \times W}, \boldsymbol{o}'_1,  \boldsymbol{o}'_2 \in \mathbb{R}^{\frac{C}{2} \times H \times W}$.

\vspace{-1mm}
Down Sample is the combination of Down Scale and Fusion Up, where we apply average pooling followed by channel concatenation. Note that we just describe the condition of double or half the size for simplicity, while it is generalized to more complex conditions with four times channels in BiDRN (see supplementary file). Redesigning these four modules allows us to bypass full-precision activations with nearly the same parameter and computational cost.

\begin{figure}
\vspace{-3mm}
 \centering
% \fbox{\parbox[c][9cm]{\linewidth}{Abstract}}
\includegraphics[width=\linewidth]{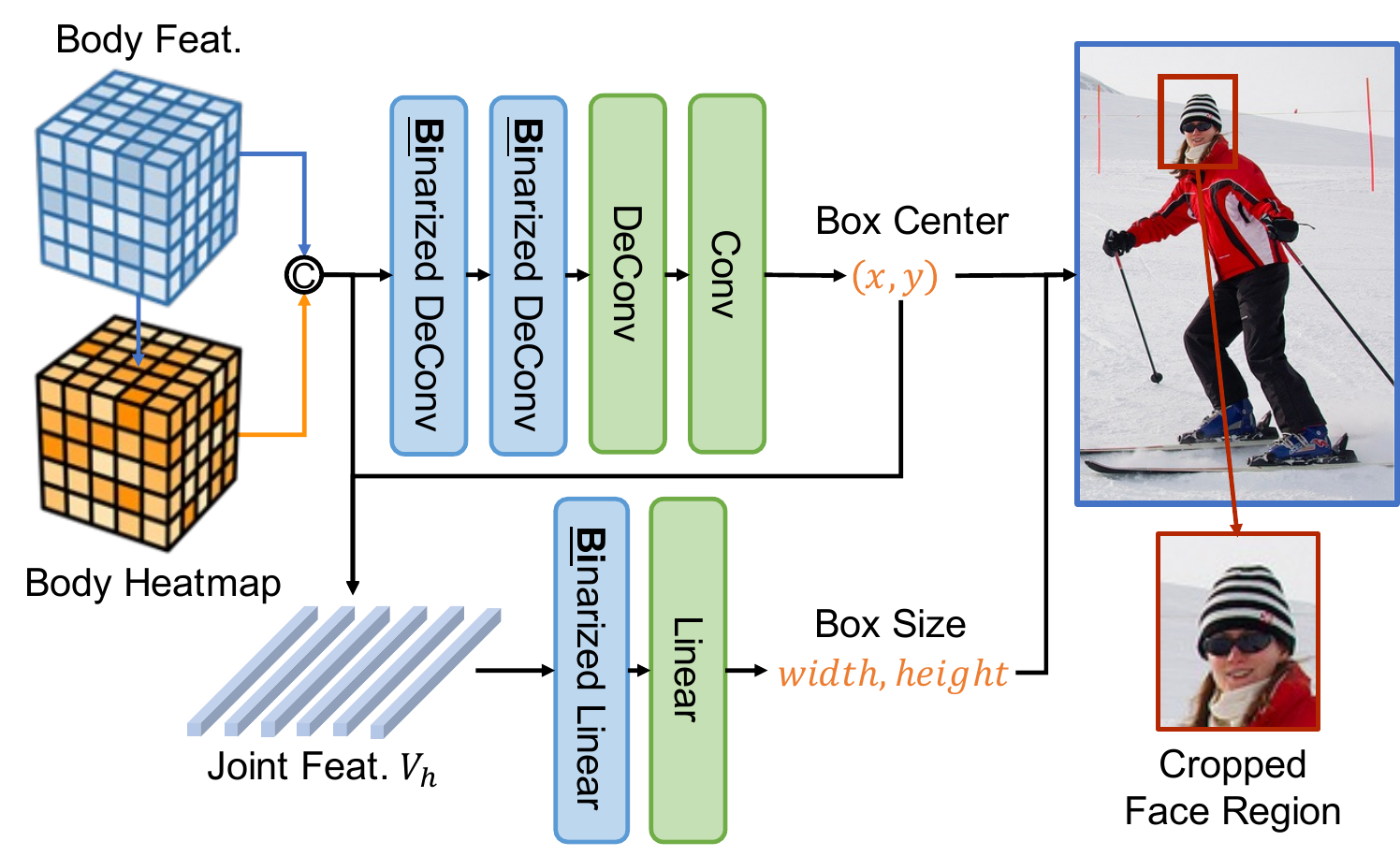}
\vspace{-8mm}
\caption{Binarized face BoxNet extracts the face region from the high-resolution human image. Hand regions are extracted by binarized hands BoxNet with the same architecture.}
\label{fig:boxnet}
\vspace{-6mm}
\end{figure}

\begin{table*}[t]
\centering
\vspace{-2.5mm}
\caption[Reconstruction errors on EHF.]{3D whole-body reconstruction error comparisons on EHF~\citep{pavlakos2019expressive} and AGORA~\citep{patel2021agora} benchmarks. $\dagger$ indicates that the model does not use pre-trained weights, as well as additional hand-only and face-only datasets for fair comparison.\label{tab:3d_smplx_results}}
% \vspace{-1mm}
\resizebox{\textwidth}{!}
{
\setlength{\tabcolsep}{1mm}
    \begin{tabular}{l|c|c|c|ccc|ccc|cc|ccc|ccc}
    \toprule
    \rowcolor{color3}
     & 
     &
     &
     & \multicolumn{8}{c|}{\boldmath{}{EHF}\unboldmath{}} & \multicolumn{6}{c}{\boldmath{}{AGORA}\unboldmath{}} \\
    \cline{5-18}
    \rowcolor{color3}
    & & & & \multicolumn{3}{c|}{\bigstrut[t] \boldmath{}{MPVPE $\downarrow$}\unboldmath{}} & \multicolumn{3}{c|}{\boldmath{}{PA-MPVPE $\downarrow$}\unboldmath{}} & \multicolumn{2}{c|}{\boldmath{}{PA-MPJPE $\downarrow$}\unboldmath{}} & \multicolumn{3}{c|}{\boldmath{}{MPVPE $\downarrow$}\unboldmath{}} & \multicolumn{3}{c}{\boldmath{}{PA-MPVPE $\downarrow$}\unboldmath{}} \\
    \cline{5-18}
    \rowcolor{color3}
    \multirow{-3}{*}{{Method}}& \multirow{-3}{*}{{Bit}} & \multirow{-3}{1.7cm}{{Params (M)}} & \multirow{-3}{1.5cm}{\hfill{OPs (G)}\hfill} & \bigstrut[t] {All} & {Hand} & {Face} & {All} & {Hand} & {Face} &{Body} & {Hand} & {All} & {Hand} & {Face} & {All} & {Hand} & {Face} \\
    \midrule
    % SMPLify-X~\citep{Pavlakos_2019smplx} & & & & & & & & \\
    ExPose & 32 & - & - & 77.1 & 51.6 & 35.0 & 54.5 & 12.8 & 5.8 & - & - & 219.8 & 115.4 & 103.5 & 88.0 & 12.1 & 4.8 \\
    % FrankMocap~\citep{Rong_2021frank} & & & & & & & & \\
    FrankMocap & 32 & - & - & 107.6 & 42.8  & - & 57.5 & 12.6 & - & -	& -  & 218.0 & 95.2 & 105.4 & 90.6 & 11.2 & 4.9 \\
    PIXIE & 32 & - & - & 89.2 & 42.8  & 32.7 & 55.0 & 11.1 & 4.6 & -	& -  & 203.0 & 89.9 & 95.4 & 82.7 & 12.8 & 5.4 \\
    Hand4Whole $\dagger$ & 32 & 77.84 & 16.85 & 86.3 & 47.2 & 26.1 & 57.5 & 13.2  & 5.8 & 70.9 &	13.3 & 194.8 & 78.6 & 88.3 & 79.0 & 9.8 & 4.8 \\
    % Hand4Whole~\citep{moon2022accurate} & 32 & & & 79.2 & 43.2 & 25.0 & 53.1 & 12.1  & 5.8 & - &	- \\
    \midrule
    BNN & 1 & 21.61 & 2.63 & 172.2 & 99.0 & 53.9 & 115.6 & 18.4 & 6.2 & 129.4 & 19.0 & 267.6 & 114.0  & 141.3 & 94.9 & 10.4 & 5.0 \\
    XNOR & 1 & 21.61 & 2.63 & 195.5 & 105.0 & 57.5 & 119.9 & 18.5 & 6.2 & 134.5 & 19.1 & 271.1 & 127.9 & 156.9 & 94.1 & 10.5 & 5.1 \\
    DoReFa & 1 & 21.61 & 2.63 & 174.4 & 93.9 & 53.9 & 109.3 & 18.4 & 6.0 & 121.3 & 19.0 & 257.6 & 115.3 & 139.4 & 93.5 & 10.4 & 5.0 \\
    Bi-Real & 1 & 21.61 & 2.63 & 161.7 & 92.7 & 48.5 & 108.7 & 18.5 & 5.9 & 121.2 & 19.1 & 242.0 & 104.3 & 121.8 & 92.6 & 10.4 & 5.0 \\
    ReActNet & 1 & 21.66 & 2.63 & 149.8 & 86.5 & 45.8 & 98.8 & 18.5 & 6.1 & 111.6 & 19.1 & 237.6 & 102.9 & 120.2 & 91.4 & 10.4 & 4.9 \\
    ReCU & 1 & 21.71 & 2.65 & 142.7 & 78.3 & 49.6 & 85.4 & 18.2 & 6.0 & 97.1 & 18.8 & 225.1 & 96.2 & 108.3 & 89.7 & 10.3 & 4.9 \\
    FDA & 1 & 32.06 & 2.81 & 171.8 & 93.7 & 53.3 & 108.5 & 18.4 & 6.1 & 120.5 & 19.0 & 256.4 & 114.6 & 138.6 & 93.0 & 10.4 & 5.0 \\
    \midrule
    \rowcolor{color4}
    \textbf{BinaryHPE} & 1 & \textbf{17.22} & \textbf{2.50} & \textbf{118.3} & \textbf{70.8} & \textbf{37.6} & \textbf{76.9} & \textbf{17.4} & \textbf{6.0} & \textbf{88.2} & \textbf{17.9} & \textbf{215.0} & \textbf{92.1} & \textbf{102.3} & \textbf{87.7} & \textbf{10.3} & \textbf{4.9} \\
    % \modelname (Ours)$\dagger$ & \textbf{81.9}&	\textbf{41.5}&	\textbf{21.2}&\textbf{42.2}&\textbf{8.6}&	\textbf{2.0}&	\textbf{48.4}&	\textbf{8.8} \\
    \bottomrule
    \end{tabular}%
    }
  % \vspace{-1mm}
  \vspace{-4.5mm}
\end{table*}

\vspace{-1mm}
\noindent\textbf{Block Residual.}\quad
Full-precision information may be diluted by binarized convolution layers, particularly in very deep networks. To address this, we propose a Block Residual mechanism that bypasses full-precision information in each block, preserving crucial details across the network.

Note that the number of blocks is significantly lower than the count of conv layers, we utilize a full-precision Conv1$\times$1 layer to extract more accurate features with acceptable parameter burden. As shown in \Cref{fig:BiDRB}, the overall BiDRB composed of both LCR and BR can be formulated as
\begin{align}
\boldsymbol{o}'' = \text{BaseLCR}(\text{DSaR}(a_f)) + \text{BR}(a_f),
\label{eq:BiDRB}
\end{align}
where BaseLCR, DSaR, and BR denote base Local Convolution Residual, Down Sample Residual, and Block Residual respectively. Note that \Cref{eq:BiDRB} is only one kind of BiDRB, other kinds of BiDRB may incorporate alternative modules such as Fusion up, Fusion Down, and Down Scale Residuals.
Moreover, a binarized version of Block Residual can be used for tasks that do not require high-quality features but require efficiency with extreme compression. 

By combining Local Convolution Residual and Block Residual, Binarized Dual Residual Block can preserve full-precision information as much as possible while maintaining nearly the same parameters and computational cost. The body, hand, and face encoders based on BiDRN can extract better image features than simple binarization methods.

\vspace{-1mm}
\subsection{Binarized BoxNet}
\vspace{-1mm}
 
The bounding boxes for the hands and face are predicted by BoxNet. Initially, it predicts 3D heatmaps of human joints $\mathbf{H}$ from the encoder output $\mathbf{F}$. These heatmaps are then concatenated with $\mathbf{F}$, and several Deconv and Conv layers are applied to this combined feature map. Afterward, soft-argmax~\citep{sun2018integral} is used to determine the box center, followed by fully connected layers to compute the box size. We observe that the parameters and computational cost of these layers, especially the Deconv layers, are significantly higher compared to other components in the decoder, which seems excessive for a few bounding box parameters.

\vspace{-1mm}
Thus, we binarize both Deconv layers and Linear layers except the final one in~\Cref{fig:boxnet}, so that we can maintain good output accuracy. Experiments (\Cref{tab:ablation_boxnet}) further show that such binarization even leads to performance gain while reduces memory and computational costs significantly.

\vspace{-1mm}
\noindent\textbf{Loss Function.}\quad
BinaryHPE is the combination of BiDRN and Binarized BoxNet. We train it end-to-end by minimizing the loss following~\citet{moon2022accurate}:
\begin{align}
    \mathcal{L} = \mathcal{L}_\text{smplx} + \mathcal{L}_\text{joint} + \mathcal{L}_\text{box},
\end{align}
where $\mathcal{L}_\text{smplx}$, $\mathcal{L}_\text{joint}$, and $\mathcal{L}_\text{box}$ are the $L_1$ distance between predicted and GT SMPL-X parameters, joint coordinates, and bounding boxes respectively.

%-------------------------------------------------------------------------

\begin{figure*}[t]
\scriptsize
% \tiny
\centering
\vspace{-1.5mm}
\begin{tabular}{ccccccccc}

% % one row
\hspace{-4.3mm}
\begin{adjustbox}{valign=t}
\begin{tabular}{cccccccccccccc}
\includegraphics[width=0.096\textwidth]{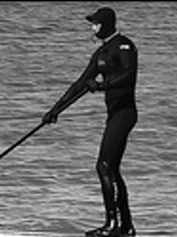} \hspace{-4.3mm} &
\includegraphics[width=0.096\textwidth]{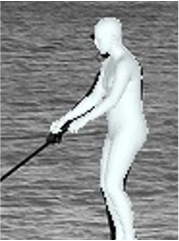} \hspace{-4.3mm} &
\includegraphics[width=0.096\textwidth]{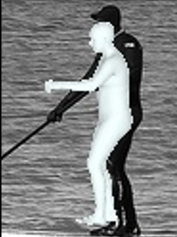} \hspace{-4.3mm} &
\includegraphics[width=0.096\textwidth]{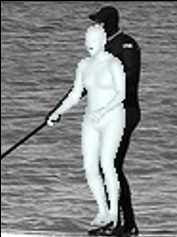} \hspace{-4.3mm} &
\includegraphics[width=0.096\textwidth]{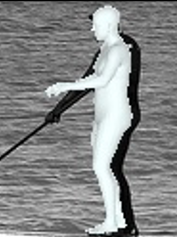} \hspace{-4.3mm} &
\includegraphics[width=0.096\textwidth]{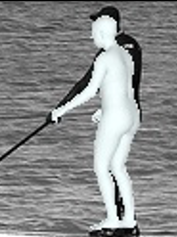} \hspace{-4.3mm} &
\includegraphics[width=0.096\textwidth]{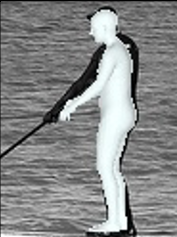} \hspace{-4.3mm} &
\includegraphics[width=0.096\textwidth]{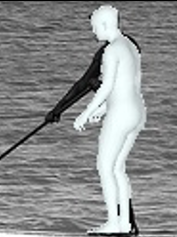} \hspace{-4.3mm} &
\includegraphics[width=0.096\textwidth]{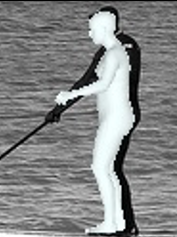} \hspace{-4.3mm} &
\includegraphics[width=0.096\textwidth]{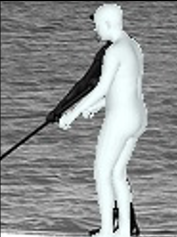} \hspace{-4.3mm} &
\\ 
\includegraphics[width=0.096\textwidth]{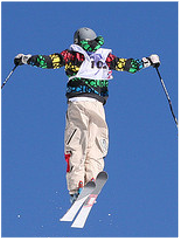} \hspace{-4.3mm} &
\includegraphics[width=0.096\textwidth]{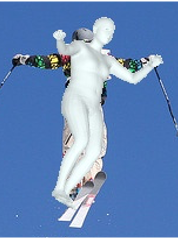} \hspace{-4.3mm} &
\includegraphics[width=0.096\textwidth]{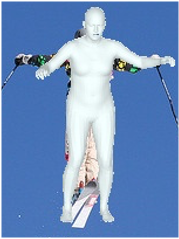} \hspace{-4.3mm} &
\includegraphics[width=0.096\textwidth]{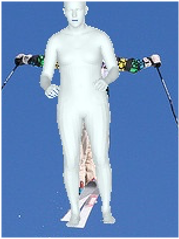} \hspace{-4.3mm} &
\includegraphics[width=0.096\textwidth]{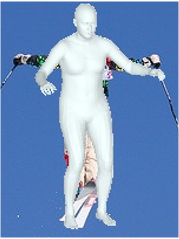} \hspace{-4.3mm} &
\includegraphics[width=0.096\textwidth]{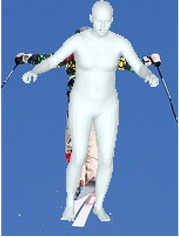} \hspace{-4.3mm} &
\includegraphics[width=0.096\textwidth]{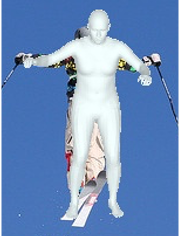} \hspace{-4.3mm} &
\includegraphics[width=0.096\textwidth]{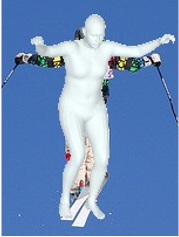} \hspace{-4.3mm} &
\includegraphics[width=0.096\textwidth]{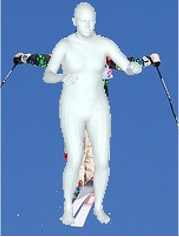} \hspace{-4.3mm} &
\includegraphics[width=0.096\textwidth]{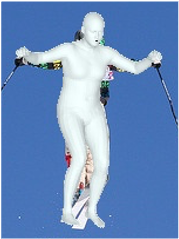} \hspace{-4.3mm} &
\\ 
\includegraphics[width=0.096\textwidth]{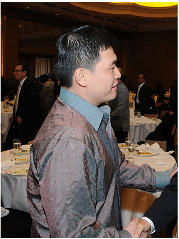} \hspace{-4.3mm} &
\includegraphics[width=0.096\textwidth]{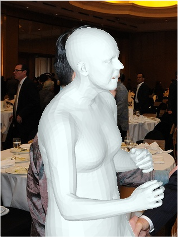} \hspace{-4.3mm} &
\includegraphics[width=0.096\textwidth]{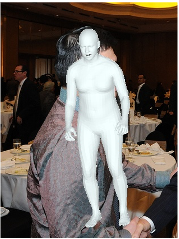} \hspace{-4.3mm} &
\includegraphics[width=0.096\textwidth]{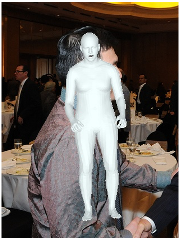} \hspace{-4.3mm} &
\includegraphics[width=0.096\textwidth]{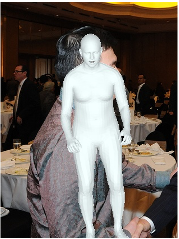} \hspace{-4.3mm} &
\includegraphics[width=0.096\textwidth]{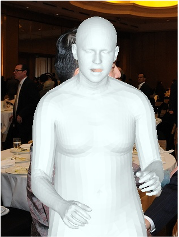} \hspace{-4.3mm} &
\includegraphics[width=0.096\textwidth]{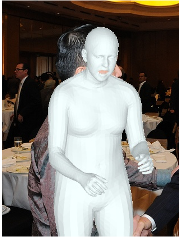} \hspace{-4.3mm} &
\includegraphics[width=0.096\textwidth]{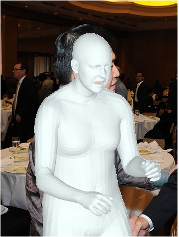} \hspace{-4.3mm} &
\includegraphics[width=0.096\textwidth]{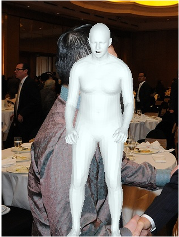} \hspace{-4.3mm} &
\includegraphics[width=0.096\textwidth]{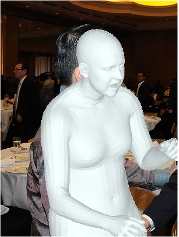} \hspace{-4.3mm} &
\\ 
\includegraphics[width=0.096\textwidth]{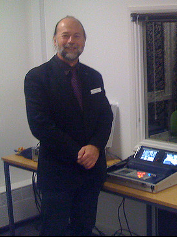} \hspace{-4.3mm} &
\includegraphics[width=0.096\textwidth]{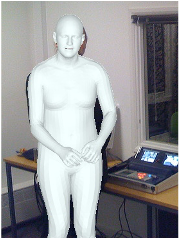} \hspace{-4.3mm} &
\includegraphics[width=0.096\textwidth]{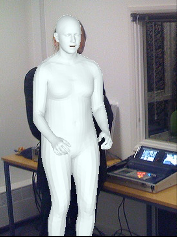} \hspace{-4.3mm} &
\includegraphics[width=0.096\textwidth]{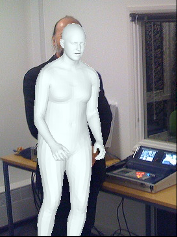} \hspace{-4.3mm} &
\includegraphics[width=0.096\textwidth]{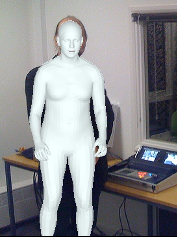} \hspace{-4.3mm} &
\includegraphics[width=0.096\textwidth]{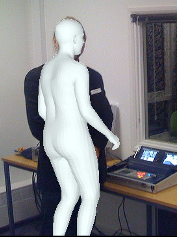} \hspace{-4.3mm} &
\includegraphics[width=0.096\textwidth]{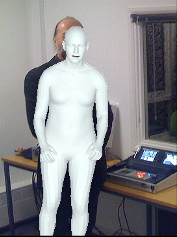} \hspace{-4.3mm} &
\includegraphics[width=0.096\textwidth]{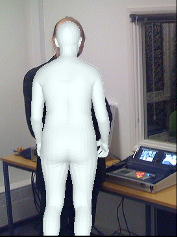} \hspace{-4.3mm} &
\includegraphics[width=0.096\textwidth]{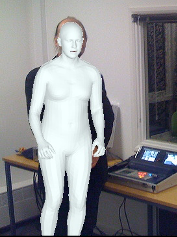} \hspace{-4.3mm} &
\includegraphics[width=0.096\textwidth]{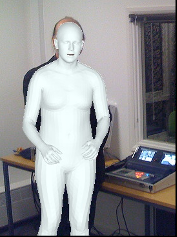} \hspace{-4.3mm} &
\\ 
\includegraphics[width=0.096\textwidth]{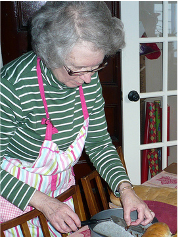} \hspace{-4.3mm} &
\includegraphics[width=0.096\textwidth]{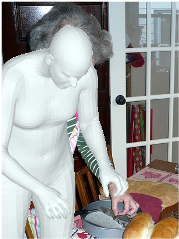} \hspace{-4.3mm} &
\includegraphics[width=0.096\textwidth]{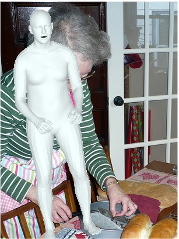} \hspace{-4.3mm} &
\includegraphics[width=0.096\textwidth]{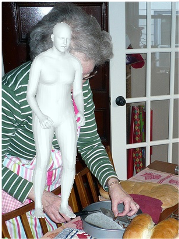} \hspace{-4.3mm} &
\includegraphics[width=0.096\textwidth]{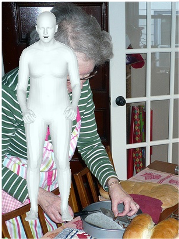} \hspace{-4.3mm} &
\includegraphics[width=0.096\textwidth]{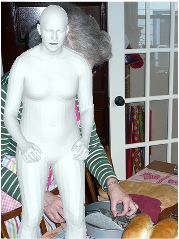} \hspace{-4.3mm} &
\includegraphics[width=0.096\textwidth]{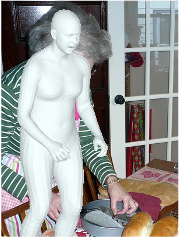} \hspace{-4.3mm} &
\includegraphics[width=0.096\textwidth]{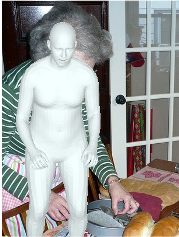} \hspace{-4.3mm} &
\includegraphics[width=0.096\textwidth]{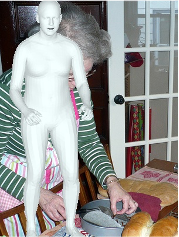} \hspace{-4.3mm} &
\includegraphics[width=0.096\textwidth]{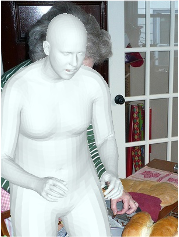} \hspace{-4.3mm} &
\\ 
Image \hspace{-4.3mm} &
Hand4Whole \hspace{-4.3mm} &
BNN \hspace{-4.3mm} &
XNOR \hspace{-4.3mm} &
DoReFa \hspace{-4.3mm} & 
Bi-Real \hspace{-4.3mm} & 
ReActNet \hspace{-4.3mm} & 
ReCU \hspace{-4.3mm} & 
FDA \hspace{-4.3mm} & 
BinaryHPE \hspace{-4.3mm} & 
\\
\end{tabular}
\end{adjustbox}

\end{tabular}
\vspace{-4.5mm}
\caption{Qualitative comparison between full-precision Hand4Whole, seven SOTA BNN-based methods, and our BinaryHPE on the MSCOCO~\citep{jin2020whole} dataset. Bypassing the full-precision information is necessary for accurate human pose and shape estimation.}
\label{fig:qualitative}
\vspace{-5mm}
\end{figure*}

\vspace{-2mm}
\section{Experiment}
\label{exp_all}
\vspace{-1mm}
\subsection{Experimental Settings}
\vspace{-1mm}
\noindent\textbf{Datasets.}\quad
We use Human3.6M~\citep{ionescu2013human3}, whole-body MSCOCO~\citep{jin2020whole} and MPII~\citep{andriluka20142d} for training. Following~\citet{moon2022accurate}, the 3D pseudo-GTs for training are obtained by NeuralAnnot~\citep{moon2022neuralannot}. To make the binarized model simple and easy to train, different from~\citet{moon2022accurate}, we do not use additional hand-only and face-only datasets, or additional stages to finetune the model. Finally, we evaluate our BinaryHPE on EHF~\citep{pavlakos2019expressive} and AGORA~\citep{patel2021agora}.

% \vspace{0.8mm}
\vspace{-1mm}
\noindent\textbf{Evaluation Metrics.}\quad
We adopt Mean Per Joint Position Error (MPJPE) and Mean Per Vertex Position Error (MPVPE), along with their aligned version PA-MPJPE and PA-MPVPE, to evaluate the performance of HPE models. Consistent with prior works~\citep{xia2022basic,qin2020forward,hubara2016binarized,cai2023binarized}, we calculate the parameters of BNN-based methods as Params = Params$_b$ + Params$_f$, where Params$_b$ = Params$_f$ / 32 represents that the binarized parameters is $1/32$ of its full-precision counterpart. Similarly, the computational complexity of BNNs is measured by operation per second (OPs), which is calculated as OPs = OPs$_b$ + OPs$_f$, where OPs$_b$ = OPs$_f$ / 64, and OPs$_f$ = FLOPs (floating point operations).

% \vspace{0.8mm}
\noindent\textbf{Implementation Details.}\quad
BinaryHPE is implemented in PyTorch~\citep{paszke2019pytorch}. To keep the pipeline concise and ensure the performance of BinaryHPE isn't due to pretraining or finetuning on additional datasets, we don't pre-train or finetune on hand-only or face-only datasets. We use Adam~\citep{kingma2014adam} optimizer with batch size 24 and initial learning rate of 1$\times$10$^{-4}$ to train BinaryHPE for 14 epochs on a single A100-80G GPU. We apply standard data augmentation techniques, including scaling, rotation, random horizontal flipping, and color jittering. We also provide a mapping table from ResNet backbones to the proposed modules of BiDRN in supplementary file.

\begin{table*}[t]
% \vspace{-1.5mm}
% subfloat c - BackBone Architecture
\vspace{-3.2mm}
\caption{Ablation study on EHF dataset evaluated by \emph{MPVPEs}, with final results in \textbf{bold}. (a) DScR, FUR, FDR, and DSaR refer to Down Scale Residual, Fusion Up Residual, Fusion Down Residual, and Down Sample Residual in~\Cref{fig:local_residual}. (d) Binarizing all networks results in \emph{MPVPEs} of $118.3$, $70.8$, $37.6$ (All, Hand, Face), compared to full-precision network's $86.3$, $47.2$, $26.1$.\vspace{0.5mm}}
% \vspace{-1mm}

\subfloat[\small Break-down ablation of LCR \label{tab:breakdown}\vspace{-1.2mm}]{
\scalebox{0.85}{\begin{tabular}{l@{\hskip 12pt}c@{\hskip 12pt}c@{\hskip 12pt}c@{\hskip 12pt}c@{\hskip 12pt}c}
\toprule
\rowcolor{color3}
Method &BaseLCR  &~+ DScR~ &~+ FUR~  &~+ FDR~ &~+ DSaR  \\
%\rowcolor{color3} &\citep{edvr} &\citep{Su} &\citep{stfan}  &\citep{tsp}  &\textbf{(Ours)} \\
				\midrule
				All MPVPEs &139.3 & 127.8 & 126.0 & 124.7 & \textbf{118.3} \\
				% Hand  &0.594    &0.737  &0.729 &0.758  &0.776 &\bf 0.837  \\
    %                 Face  &0.594    &0.737  &0.729 &0.758  &0.776 &\bf 0.837  \\
                    Params (M) & 17.05 & 17.05 & 17.14 & 17.21 & 17.22 \\
				OPs (G) & 2.48 & 2.48 & 2.49 & 2.50 & 2.50 \\
				\bottomrule
\end{tabular}}}\hfill
\subfloat[\small Study of pre-activation function\label{tab:pre-activation}\vspace{-1.2mm}]{
		\scalebox{0.85}{
			\begin{tabular}{l@{\hskip 12pt}c@{\hskip 12pt}c@{\hskip 12pt}c@{\hskip 12pt}c}
				\toprule
				\rowcolor{color3}Activation Type &Additional Params &All &Hand &Face   \\
				\midrule
				Hardtanh($x_f$)   & No & \textbf{118.3} & \textbf{70.8} & \textbf{37.6} \\
				ReLU($x_f$)   & No & 126.8 & 71.5 & 38.9 \\
				PReLU($x_f$)  & Yes & 125.9 & 70.6 & 37.3  \\
				\bottomrule
	\end{tabular}}}\vspace{1mm}\\
	% subfloat a - RoIAlign (ResNet-50-C5)
\subfloat[\small Ablation study of Block Residual (BR) \label{tab:br}\vspace{-1.2mm}]{ 
		\scalebox{0.85}{
			\begin{tabular}{l@{\hskip 8pt}c@{\hskip 8pt}c@{\hskip 8pt}c@{\hskip 12pt}c@{\hskip 12pt}c}
				%\small
				\toprule
				\rowcolor{color3}Method &Params (M) & OPs (G) &All &Hand &Face \\
				\midrule
				w/o BR & 11.51 & 1.25 & 139.6 & 85.4 & 39.1 \\
				Binarized BR & 11.68 & 1.28 & 120.0 & 73.3 & 37.9 \\
				Full-precision BR & 17.22 & 2.50 & \textbf{118.3} & \textbf{70.8} & \textbf{37.6} \\
				% \checkmark  &\checkmark &\checkmark &\bf 29.76 &\bf 0.837  \\
				\bottomrule
	\end{tabular}}}\hfill
	% subfloat d - Multinomial vs Independent Masks
	% subfloat b - mask representation
\subfloat[\small Ablation study of binarizing different parts \label{tab:part}\vspace{-1.2mm}]{
		\scalebox{0.85}{
			\begin{tabular}{l@{\hskip 8pt}c@{\hskip 8pt}c@{\hskip 8pt}c@{\hskip 12pt}c@{\hskip 12pt}c}
				\toprule
				\rowcolor{color3}Binarized Network &Params (M) &OPs (G) &All &Hand &Face\\
				\midrule
				Body Encoder & 47.78 & 7.45 &119.8 &65.9 &36.7 \\
				Hand Encoder & 47.78 & 7.45 &86.0 &49.0 &27.9 \\
				Face Encoder & 57.08 & 9.94 &86.8 &55.3 &25.9 \\
                    % Body \& Hand \& Face & & &122.6 &68.6 &39.6 \\
				\bottomrule
                \end{tabular}}}
\label{tab:ablations}
 \vspace{-4mm}
% \vspace{-7mm}
\end{table*}

\vspace{-2mm}
\subsection{Quantitative Results}
\vspace{-1mm}
We compare BinaryHPE with 7 SOTA BNN-based methods: BNN~\citep{hubara2016binarized}, XNOR~\citep{rastegari2016xnor}, DoReFa~\citep{zhou2016dorefa}, Bi-Real~\citep{liu2018bi}, ReActNet~\citep{liu2020reactnet}, ReCU~\citep{xu2021recu}, and FDA~\citep{xu2021learning}. To adapt these BNN methods to the HPE task, we replace the convolutional layers in Hand4Whole's ResNet backbone with binary convolutions from the corresponding BNN. The rest of the model remains unchanged, following the convention for model binarization.
Besides, we also compare it with 4 SOTA 32-bit full-precision methods, including ExPose~\citep{choutas2020monocular}, FrankMocap~\citep{rong2021frankmocap}, PIXIE~\citep{feng2021collaborative}, and Hand4Whole~\citep{moon2022accurate}.

\vspace{-1mm}
\Cref{tab:3d_smplx_results} presents the performance comparisons on both EHF and AGORA datasets. It can be observed that although existing SOTA BNN-based methods can compress the model to only 27.8\% (21.61/77.84) of the original Params and 15.6\% (2.63/16.85) of the original OPs, directly applying them to the HPE task achieves poor performance. In comparison, our BinaryHPE achieves superior performance compared to these SOTA BNN-based methods with even fewer parameters and operations demands. Specifically, the \emph{All MPVPEs} of BinaryHPE show $31.3$\%, $39.5$\%, $32.2$\%, $26.8$\%, $21.0$\%, $17.1$\%, and $31.1$\% improvements 
than BNN, XNOR, DoReFa, Bi-Real, ReActNet, ReCU, and FDA on EHF dataset respectively. Furthermore, the AGORA dataset results reinforce the strengths of BinaryHPE. As shown in the right half of~\Cref{tab:3d_smplx_results}, BinaryHPE outperforms 7 SOTA BNN-based methods. Compared to the basic BNN, our BinaryHPE improves \emph{MPVPEs} by $19.7$\%, $19.2$\%, and $27.6$\% for body, hands, and face, respectively.

\vspace{-1mm}
When compared to the 32-bit full-precision methods, BinaryHPE also achieves comparable performance with extremely lower memory and computational cost. For the EHF dataset, BinaryHPE impressively narrows the \emph{All MPVPEs} gap between full-precision Hand4Whole and binarization methods from \textbf{85.9} to just \textbf{32.0}. For the AGORA dataset, surprisingly, BinaryHPE even surpasses full-precision frameworks ExPose and FrankMocap. Given that AGORA is a more complex and natural dataset~\citep{moon2022accurate,patel2021agora}, it can better demonstrate the effectiveness of BinaryHPE. This also suggests that it will be more valuable to binarize a powerful model (\textit{e.g.,} Hand4Whole), as it may perform better even after a lightweight adaptation.

\begin{figure}
% \small
\scriptsize
\centering
\vspace{-0.6mm}
% \resizebox{0.50\textwidth}{!}{
% % one row

\begin{adjustbox}{valign=t}
\hspace{-3mm}
\begin{tabular}{cccc}
\includegraphics[width=0.245\linewidth]{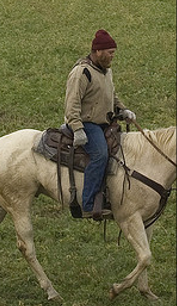} \hspace{-4mm} &
\includegraphics[width=0.245\linewidth]{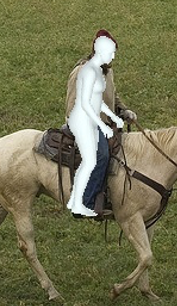} \hspace{-4mm} &
\includegraphics[width=0.245\linewidth]{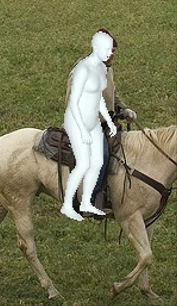} \hspace{-4mm} &
\includegraphics[width=0.245\linewidth]
{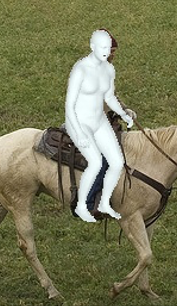} \hspace{-4mm}
\\
% \multirow{2}{*}{Image}  \hspace{-4mm} &
Image \hspace{-4mm} &
w/o BR  \hspace{-4mm} &
Binarized BR \hspace{-4mm} &
Full-precision BR \hspace{-4mm}
\end{tabular}
\end{adjustbox}
%}
\vspace{-2.5mm}
\caption{\small{Visual comparison of Block Residual ablation study.}}
\label{fig:ablation_3}
\vspace{-5.9mm}
\end{figure}

\vspace{-2mm}
\subsection{Qualitative Results}
\vspace{-1mm}
Following~\citet{moon2022accurate} and~\citet{lin2023one}, we show the qualitative results on MSCOCO dataset in~\Cref{fig:qualitative}. It can be observed that the 3D human meshes recovered by previous BNN methods cannot even match the 2D images, resulting in completely incorrect results. Conversely, BinaryHPE consistently aligns with 2D images, even in complex backgrounds, as highlighted in the third and fifth rows. Moreover, previous BNN approaches tend to generate wrong rotations, \eg, the second and fourth rows. While BinaryHPE keeps the original rotations, as well as capturing more accurate facial expressions and hand poses. Finally, BinaryHPE exemplifies greater stability compared to traditional BNNs, achieving accurate and consistent estimations across all images. More visual comparisons of EHF and AGORA datasets are shown in supplementary file.

\vspace{-2mm}
\subsection{Ablation Study}
\vspace{-1mm}
\noindent\textbf{Break-down Ablation.}\quad
We first establish a baseline with base Local Convolution Residual (LCR). Next, we incrementally introduce Down Scale Residual (DScR), Fusion Up Residual (FUR), Fusion Down Residual (FDR), and Down Sample Residual (DSaR) to improve performance. Notably, our baseline LCR (BaseLCR) achieves an \emph{All MPVPEs} of 139.3, already outperforming the basic BNN (172.2) by a significant margin. As shown in~\Cref{tab:breakdown}, when we successively use DScR, FUR, FDR, and DSaR, the \emph{All MPVPEs} is reduced by 11.5, 1.8, 1.3, and 6.4 respectively. They together reduce the \emph{All MPVPEs} by 21.0 in total with just a few additional Params and OPs, demonstrating the effectiveness of LCR and its four derived modules.

% \vspace{0.8mm}
\vspace{-1mm}
\noindent\textbf{Pre-activation.}\quad
We compare the Hardtanh pre-activation used in BinaryHPE with the previous default pre-activation functions ReLU and PReLU. As shown in~\Cref{tab:pre-activation}, when replacing ReLU or PReLU with Hardtanh, the \emph{All MPVPEs} can be reduced by 8.5 and 7.6 respectively without additional parameters. This suggests the superiority of the Hardtanh pre-activation in BinaryHPE.

% \vspace{0.8mm}
\vspace{-1mm}
\noindent\textbf{Block Residual.}\quad
To study the effect of Block Residual, we remove it from BinaryHPE, and also compare it with its binarization counterpart. As shown in~\Cref{tab:br}, without Block Residual, our method can still achieve 139.6 \emph{All MPVPEs}, outperforming basic BNN (172.2) with only half of the Params and OPs. Adding the binarized BR reduces the \emph{All MPVPEs} by 19.6, a significant improvement with minimal increase in Params and OPs. Replacing the binarized BR with full-precision BR further reduces the \emph{All MPVPEs} by 1.7.
Although the quantitative improvement from full-precision is modest, the qualitative results in~\Cref{fig:ablation_3} highlight its importance for accurate human pose and shape estimation. Full-precision BR recovers the hand position and rotation accurately, while Binarized BR performs well on the body but misaligns the hands.

% \vspace{0.8mm}
\vspace{-1mm}
\noindent\textbf{Binarizing Different Networks.}\quad
Since the body, hand, and face have separate encoders, we binarize them individually while keeping the others full-precision to evaluate their impact. The results in \Cref{tab:part} reveal:
\textbf{(1)} Binarizing an encoder generally leads to a performance drop, except for the Face Encoder, where \emph{MPVPE} improves. This suggests the full-precision face encoder has redundancies, and our binarization method effectively retains key full-precision information.
\textbf{(2)} Binarizing the Body Encoder also degrades hand and face performance, while binarizing the Hand or Face Encoder has little effect on others. This indicates the Body Encoder plays a central role in HPE, as it not only extracts body features but also predicts face and hand bounding boxes. Thus, preserving full-precision information in the Body Encoder is vital for overall performance.

\begin{table}
\vspace{-4mm}
\caption{{Abalation study of BoxNet on EHF, where both binarized and full-precision BoxNets are trained with Binarized BR.}\label{tab:ablation_boxnet}}
\centering
% \vspace{-2mm}
\scalebox{0.83}{
\hspace{-2mm}
\begin{tabular}{l@{\hskip 2pt}c@{\hskip 6pt}c@{\hskip 6pt}c@{\hskip 4pt}c@{\hskip 3pt}c}
				\toprule
				\rowcolor{color3}Method~ &Params (M) &OPs (G) &~All~ &~Hand~ &~Face~\\
				\midrule
				Full-precision BoxNet & 21.81 & 1.87 & 130.7 & 78.4 & 40.5 \\
                    Binarized BoxNet & 11.68 & 1.28 & 125.5 & 75.1 & 39.0 \\
                    % Body \& Hand \& Face & & &122.6 &68.6 &39.6 \\
				\bottomrule
\end{tabular}
}
\vspace{-5mm}
\end{table}

% \vspace{0.8mm}
\vspace{-1mm}
\noindent\textbf{BoxNet.}\quad
To further verify the effectiveness of Binarized BoxNet, we compare it with the full-precision BoxNet. As shown in~\Cref{tab:ablation_boxnet}, Binarized BoxNet achieves even better performance with much fewer parameters and operations, suggesting that the full-precision BoxNet is redundant and will lead to a performance drop.

\vspace{-2mm}
\section{Conclusion}
\vspace{-1mm}
In this work, we propose BinaryHPE, a novel BNN-based method for 3D human pose and shape estimation. To the best of our knowledge, this is the first work to study the binarization of the HPE task. The key to preserving estimation accuracy is to maintain the full-precision information as much as possible. To this end, we present a new binarized backbone BiDRN with Local Convolution Residual and Block Residual. Comprehensive quantitative and qualitative experiments demonstrate that BinaryHPE significantly outperforms SOTA BNNs and even achieves comparable performance with full-precision HPE methods.

% \newpage

\section*{Impact Statement}

3D human pose and shape estimation is a fundamental technology for understanding human behavior, with broad applications in AR/VR~\citep{wang2021scene}, sign language recognition~\citep{camgoz2020multi}, and emotion recognition~\citep{lin2023one}. As the volume of images and videos continues to grow, efficiently recovering 3D whole-body human meshes becomes increasingly important. Our BinaryHPE achieves more efficient and accurate 3D mesh estimations compared to existing SOTA BNN-based methods, and we hope this work inspires further research in efficient 3D human pose and shape estimation.

At present, neither 3D human pose and shape estimation techniques nor the proposed BinaryHPE exhibit any foreseeable negative societal impacts.

\nocite{langley00}

\bibliography{Reference}
\bibliographystyle{icml2025}

%%%%%%%%%%%%%%%%%%%%%%%%%%%%%%%%%%%%%%%%%%%%%%%%%%%%%%%%%%%%%%%%%%%%%%%%%%%%%%%
%%%%%%%%%%%%%%%%%%%%%%%%%%%%%%%%%%%%%%%%%%%%%%%%%%%%%%%%%%%%%%%%%%%%%%%%%%%%%%%
% APPENDIX
%%%%%%%%%%%%%%%%%%%%%%%%%%%%%%%%%%%%%%%%%%%%%%%%%%%%%%%%%%%%%%%%%%%%%%%%%%%%%%%
%%%%%%%%%%%%%%%%%%%%%%%%%%%%%%%%%%%%%%%%%%%%%%%%%%%%%%%%%%%%%%%%%%%%%%%%%%%%%%%
\newpage
\appendix
\onecolumn

% You can have as much text here as you want. The main body must be at most $8$ pages long.
% For the final version, one more page can be added.
% If you want, you can use an appendix like this one.  

% The $\mathtt{\backslash onecolumn}$ command above can be kept in place if you prefer a one-column appendix, or can be removed if you prefer a two-column appendix.  Apart from this possible change, the style (font size, spacing, margins, page numbering, etc.) should be kept the same as the main body.
%%%%%%%%%%%%%%%%%%%%%%%%%%%%%%%%%%%%%%%%%%%%%%%%%%%%%%%%%%%%%%%%%%%%%%%%%%%%%%%
%%%%%%%%%%%%%%%%%%%%%%%%%%%%%%%%%%%%%%%%%%%%%%%%%%%%%%%%%%%%%%%%%%%%%%%%%%%%%%%

\end{document}